\documentclass{article}

\usepackage[utf8]{inputenc} 
\usepackage[T1]{fontenc}    
\usepackage{hyperref}       
\usepackage{url}            
\usepackage{booktabs}       
\usepackage{amsfonts}       
\usepackage{nicefrac}       
\usepackage{microtype}      

\usepackage{graphicx}
\usepackage{amsmath}
\usepackage{booktabs}
\usepackage{multicol}

\title{\LARGE \bf
Defending Malware Classification Networks Against Adversarial Perturbations with Non-Negative Weight Restrictions
}

\author{Alex Kouzemtchenko \\ Google \\ kuza55@gmail.com}

\begin{document}

\maketitle
\thispagestyle{empty}
\pagestyle{empty}

\begin{abstract}

There is a growing body of literature showing that deep neural networks are vulnerable to adversarial input modification. Recently this work has been extended from image classification to malware classification over boolean features. In this paper we present several new methods for training restricted networks in this specific domain that are highly effective at preventing adversarial perturbations. We start with a fully adversarially resistant neural network that has hard non-negative weight restrictions and is equivalent to learning a monotonic boolean function and then attempt to relax the constraints to improve classifier accuracy.

\end{abstract}

\section{Introduction}

Grosse et al\cite{Grosse2016AdversarialClassification} showed that adversarial perturbation techniques pioneered on images can be applied to the more sensitive, adversarial application domain of malware classification. Unlike prior work on images that used a small pixel distance as a proxy for no semantic changes, their work utilizes the insight that adding functionality to a malware sample should not cause a classifier to believe it to be benign. In that light they use gradient-based techniques to determine which boolean features can be enabled to convince a malware classifier to treat a malicious sample as benign. This technique results in the majority of samples being misclassified in less than 20 steps.

In this paper we replicate their experiments and evaluate several restrictions based on non-negative weight constraints that can make networks fully resistant to this perturbation method and evaluate approaches with softer restrictions in an attempt to recover the full original accuracy.

\section{Background}

\subsection{DREBIN Dataset}

The malware classifiers in this paper are all trained on the DREBIN Android malware dataset\cite{Arp2014DREBIN:Pocket.}. It consists of 129,013 Android applications including 5,560 malicious applications. The features consist of 545,333 sparse binary features with an mean density of 48 features per application. The features are further divided into two groups: ~300,000 features derived from applications' AndroidManifest.xml file and ~200,000 features derived from runtime analysis\cite{Spreitzenbarth2013Mobile-sandbox:Applications} of the applications' behaviors.

\subsection{Network Architecture}

The architecture used by Grosse et al\cite{Grosse2016AdversarialClassification} that we will be examining is a fully connected multi-layer network with two hidden layers, ReLU activations and 50\% Dropout during training; Grosse et al found they had best results with an architecture that had 200 neurons in each layer and trained the network on 1000 sample mini-batches with a malware ratio of 0.3, i.e. 300 malware samples per batch. We implement this architecture in Keras\cite{Chollet2015Keras} and have uploaded the code to Github.

\subsection{Adversarial Peturbation Method}

To make these attacks clearly practical Grosse et al restrict themselves to only modifying features that were derived from the AndroidManifest.xml file and only allow themselves to enable these binary features.

With these restrictions, the method for modifying malware samples is:

\begin{itemize}

\item Differentiate the manifest inputs with respect to the desired class label
\item Find the feature with the greatest negative gradient that satisfies the restrictions and enable it
\item Iterate until the malware classifier is fooled or a cut off is reached

\end{itemize}

\section{Non-Negative Weight Restrictions}

The core idea in this paper is that in a domain consisting of binary features, finding adversarial perturbations is similar to finding rules in a traditional malware classification system of the form
\begin{math} A \land B \land \neg C \Rightarrow Malware \end{math}
 and then enabling the C feature in our malware sample.

As such we would like to constrain our network so that it is unable to learn such rules. In the context of Neural Networks this can be encoded by having a single output neuron and restricting the network's ability to learn negative weights.

\subsection{Hard Constraints}

Taking this idea to it's logical extreme, we can place hard restrictions on all of the weights and then train the network as usual. This is perfectly effective at preventing adversarial perturbations in theory and in practice, with the main outstanding question being whether the network can still learn to perform it's intended function. Previous work on Non-Negative MNIST\cite{Chorowski2015LearningConstraints} suggested that this is not an entirely futile effort.

This proved to be surprisingly effective; training the above network with hard non-negative weight restrictions for 30 epochs resulted in a 0.75\% false positive rate and 6.29\% false negative rate, which seems far better in terms of accuracy than all of the approaches in the original paper. However, the fact that it cannot learn these rules even when they may be highly useful seems suboptimal.

\subsection{N1/N2 Regularization}

In an attempt to restrict the network from learning problematic rules, but still allow the optimizer some freedom we defined N1/N2 loss penalties as the L1/L2 penalties applied to negative weight values:

\begin{multicols}{2}
  \[
    N1(x) = \begin{cases} 
      0 & x\geq 0 \\
      abs(x) & x\leq 0 
   \end{cases}
  \]
  \break
  \[
    N2(x) = \begin{cases} 
      0 & x\geq 0 \\
      x^2 & x\leq 0 
   \end{cases}
  \]
\end{multicols}

In an attempt to investigate the impact and correct scale, we ran a grid search on combinations of N1/N2 weight penalties, with the results presented in a heatmap in the appendix.

The best network that resulted from these was trained with a 0.67 N1 weight and no N2 regularization, which somewhat surprisingly did worse than the hard restricted network across all metrics. N2 regularization did not achieve any meaningful misclassification resistance.

\subsection{Alternative Regularization placement}

One theory for the worse results from regularization was that even a small negative weight could be later multiplied by a large positive value to become a large negative value, and a weight regularizer would not penalize this.

Two strategies were tried to deal with this, the first was regularizing activation values. The second was to introduce penalties on activations of the "pre-sum" values of matrix multiplications. I.e. when a normal matrix multiplication would be defined as on the left, we define an N1 regularization penalty as shown on the right:

\begin{multicols}{2}
  \[
    (xW)_{ij} = \sum_{k=1}^{m} A_{ik}B_{kj}
  \]
  \break
  \[
    loss = \sum_{k=1}^{m} N1(A_{ik}B_{kj})
  \]
\end{multicols}

Neither of these alternative placements significantly improved results.

\subsection{Non-negative initialization}

Unless otherwise noted, the results in this paper were trained using Glorot Normal initialization, however this initialization starts with half the weights below zero, leading the optimization process to spend a significant amount of time simply reducing the regularization term. Initializing weights with the absolute value of the Glorot initialization helped improve convergence time on these networks significantly, but did not significantly improve performance.

\section{Results}

While this paper attempts to replicate the results from Grosse et al\cite{Papernot2015DistillationNetworks} as closely as possible, the baseline results we achieved in terms of both false positives and false negatives are significantly better despite no attempts at tuning, so our false positive and false negative ratios are not directly comparable. Our baseline misclassification rates, however, are far worse. In the hopes of making these more easily comparable we implemented the distillation defence since it was the most successful in the original paper.

The distillation results from our replication are also significantly better. The main difference seems to be that in their work Grosse et al had issues with classification performance becoming much worse with a distillation temperature as low as 10, however our implementation still provided good classification at temperatures as high as 66 or 100, however this result turns out to be due to gradient masking rather than resistance to these inputs as discussed later.

\renewcommand{\thefootnote}{\fnsymbol{footnote}}
\begin{table}[h]
\begin{minipage}{\textwidth}
\centering
\label{my-label}
\caption{Experimental Results}
\begin{tabular}{@{}llllll@{}}
\toprule
Experiments from Grosse16\cite{Grosse2016AdversarialClassification}      & MWR & Epochs & MR\footnote{Misclassification Rate}      & FNR     & FPR    \\
\midrule
Unhardened Network               & 0.3 & 10     & 63.08\% & 9.73\%  & 1.29\% \\
Distillation, Temperature < 10   & 0.3 & 10     & 57\%    & 15\%    & 1\%    \\
Distillation, Temperature < 10   & 0.4 & 10     & 37\%    & 14\%    & 2\%    \\
Distillation, Temperature < 10   & 0.5 & 10     & 35\%    & 14\%    & 4\%    \\
\midrule
Replication Results              & MWR & Epochs & MR      & FNR     & FPR    \\
\midrule
Unhardened Network               & 0.3 & 10     & 96.7\%  & 6.29\%  & 0.21\% \\
Distillation, Temperature 100    & 0.3 & 10     &  3.4\%  & 5.4\%   & 0.5\% \\
Distillation, Temperature 66     & 0.4 & 10     &  5.2\%  & 5.8\%   & 0.70\% \\
Distillation, Temperature 66     & 0.5 & 10     &  6.8\%  & 4.3\%   & 1.05\% \\
\midrule
Our Results                      & MWR & Epochs & MR      & FNR     & FPR    \\
\midrule
Hard Restricted Network          & 0.3 & 30     & 0\%     & 6.29\%  & 0.75\% \\
Manifest-Restricted Network      & 0.3 & 10     & 0\%     &         &        \\
0.67 N1 Regularization           & 0.3 & 30     & 18.86\% & 11.15\% & 1.71\% \\
Fallback Network                 & 0.3 & 20     &         &         &        \\
\bottomrule
\end{tabular}
\end{minipage}
\end{table}

Heat maps of our grid searches of the various distillation parameters \& N1/N2 multipliers can be seen in the appendix.

\section{Discussion}

It it interesting to observe that the hard restricted network is equivalent to a monotonic boolean function, and it may be fruitful to utilize them to train hard-restricted models, though this has not been explored.

Recent work\cite{Papernot2016TowardsLearning} has also shown that defensive distillation may be masking the gradients of the model, rather than actually preventing adversarial perturbations. I have replicated this result showing that if you take the adversarial examples generated using the unhardened network 95.3\% of them are misclassified by the network trained using defensive distillation at temperature 100.

\section{Conclusions}

In this paper we showed that it is possible to modify the network designed by Grosse et al to be fully resistant to the adversarial perturbation process they designed without significant loss of accuracy.

We have also shown that distillation can achieve good naive resistance to adversarial examples in this domain, though like other work has shown, this resulting network is only resistant due to gradient masking and these perturbations are still effective when found through a proxy network.

\ 

\ 

\ 

\ 

\ 

\ 

\section{Appendix}

\begin{figure*}[ht]
  \includegraphics[width=\textwidth,height=7cm]{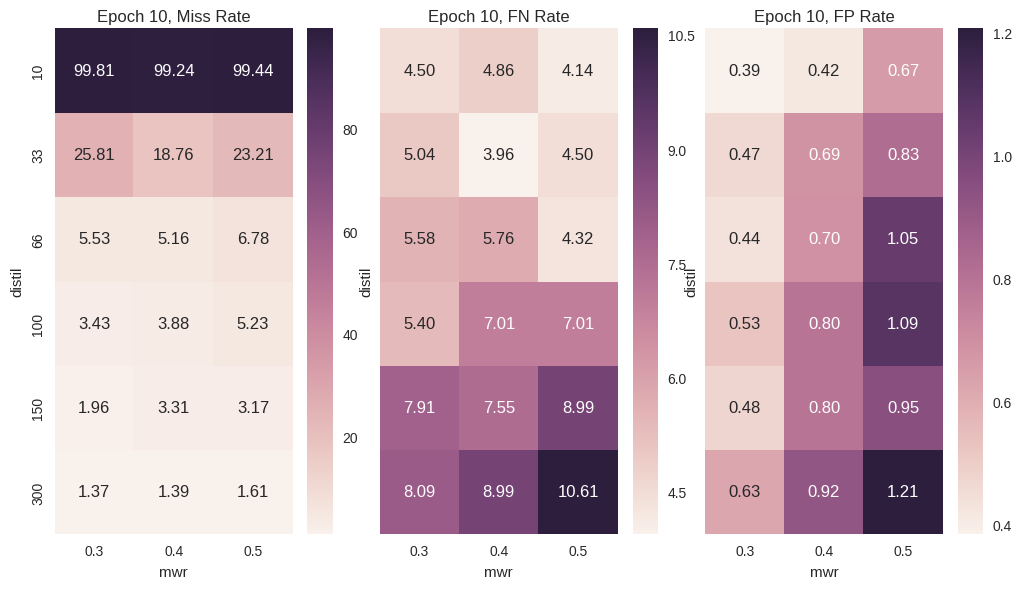}
  \caption{Results of Distillation Experiments}
\end{figure*}

\begin{figure*}[ht]
  \includegraphics[width=\textwidth,height=7cm]{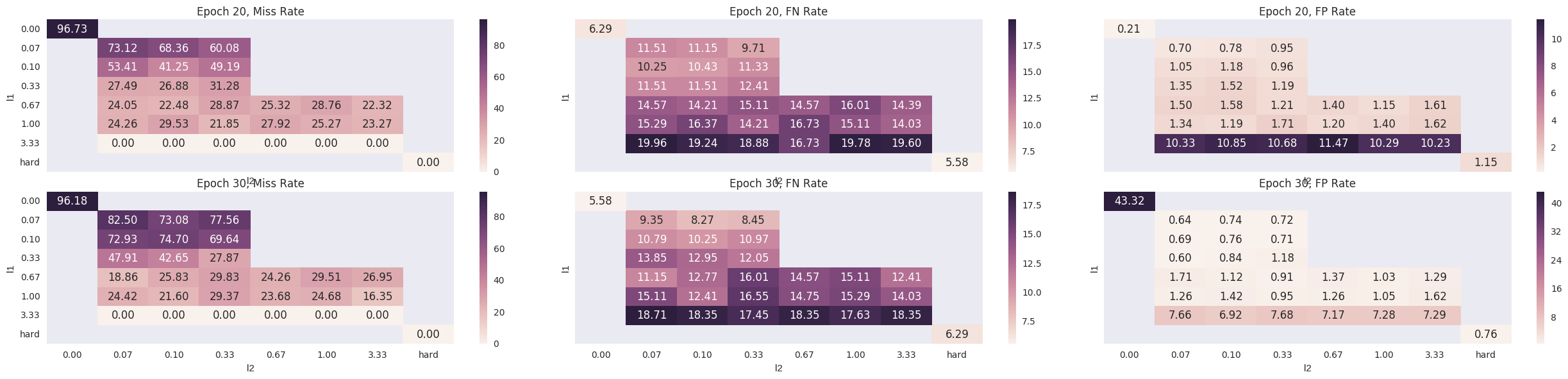}
  \caption{Results of N1/N2 Regularization Experiments with MWR@0.3}
\end{figure*}

\bibliography{Mendeley.bib}{}
\bibliographystyle{IEEEtran}

\end{document}